\documentclass[letterpaper, 10 pt, conference]{ieeeconf}  

\IEEEoverridecommandlockouts                              

\usepackage{cite}
\usepackage{amsmath,amssymb,amsfonts}
\usepackage{algorithmic}
\usepackage{graphicx}
\usepackage{textcomp}
\usepackage{xcolor}
\usepackage{booktabs}
\usepackage{tabularx}
\def\BibTeX{{\rm B\kern-.05em{\sc i\kern-.025em b}\kern-.08em
    T\kern-.1667em\lower.7ex\hbox{E}\kern-.125emX}}

\usepackage{array}
\usepackage[caption=false,font=normalsize,labelfont=sf,textfont=sf]{subfig}
\usepackage{stfloats}
\usepackage{url}
\usepackage{verbatim}
\usepackage{cite}
\usepackage[utf8]{inputenc}
\usepackage{caption}
\graphicspath{ {./Figures/} }

\begin{document}

\title{\LARGE \bf
Improving Drone Imagery For Computer Vision/Machine Learning in Wilderness Search and Rescue}

\author{Robin Murphy and Thomas Manzini
\thanks{R. Murphy and T. Manzini are with Texas A\&M University, College Station, TX, USA, and the Center for Robot-Assisted Search and Rescue}}

\maketitle
\thispagestyle{empty}
\pagestyle{empty}

\begin{abstract}
    This paper describes gaps in acquisition of drone imagery that impair the use with computer vision/machine learning (CV/ML) models and makes five recommendations to maximize image suitability for CV/ML post-processing. 
    It describes a notional work process for the use of drones in wilderness search and rescue incidents. The large volume of
    data from the wide area search phase offers the greatest opportunity for CV/ML techniques because of the large number of images that would otherwise have to be manually inspected. The 2023 Wu-Murad search in Japan, one of the largest missing person searches conducted in that area, serves as a case study. 
    Although drone teams conducting wide area searches may not know in advance if the data they collect is going to be used for CV/ML post-processing, there are data collection procedures that can improve the search in general with automated collection software. If the drone teams do expect to use CV/ML, then they can exploit knowledge about the model to further optimize flights.
The paper identifies opportunities for 
AI path planning for image acquisition, including managing altitude and adaption to mountainous terrain, 
multirobot coordination and tasking, and
platform design, especially in terms of choice of sensors and software functionality. 
\end{abstract}

\section{Introduction}

Small uncrewed aerial systems (sUAS), commonly referred to as drones, have been credited for saving 1002 lives in 600 incidents in 39 countries \cite{dji}. However, there is a surprising lack of application of computer vision/machine learning (CV/ML) methods for finding missing persons in wilderness search and rescue (WiSAR). 
While numerous papers, too many to cite, 
discuss the potential for CV/ML, only three incidents appear have used it for a real incident. Of those three, only one---a 2021 deployment by a team in Poland using the SARUAV system---successfully identified the missing person \cite{Niedzielski:2021,Tunsio:2022}.
Two other attempts, the
2015 Blanco River Texas floods \cite{proft:2015} and the 
2023 Wu-Murad missing person in Japan \cite{manzini:ICCV}, did not find the persons.  

The small number of reports raises the question: Why is CV/ML not more effective?
Based on the Center for Robot-Assisted Search and Rescue's engagement in the recent Wu-Murad search, there appear to be two culprits:
\begin{itemize}
\item Non-representative models: CV/ML models are based on datasets that 
do not represent missing persons and their behavior
do not include diversity of terrains, ground cover, and weather
\item Unsuitable acquisition of imagery: Drone teams are not collecting imagery in a manner conducive for CV/ML post-processing
\end{itemize}

The issue of non-representative CV/ML and datasets is discussed in \cite{manzini:ICCV}. 
This paper will detail the gaps in acquisition using the Wu-Murad search as a case study. It will first describe the notional use of drones for searching for missing persons in the wilderness. Next it will summarize the Wu-Marad missing person search, the largest such search in that area of Japan \cite{nbcCT}. It will then make recommendations for the AI path planning, 
multirobot coordination and tasking, and 
platform design communities as to how to build and program
 drone flights to maximize acquisition of images suitable for CV/ML post-processing.

 \section{Notional Work Process for Drones at WiSAR Incidents}

 \begin{figure*}[htbp]
\centerline{\includegraphics[width=\textwidth]{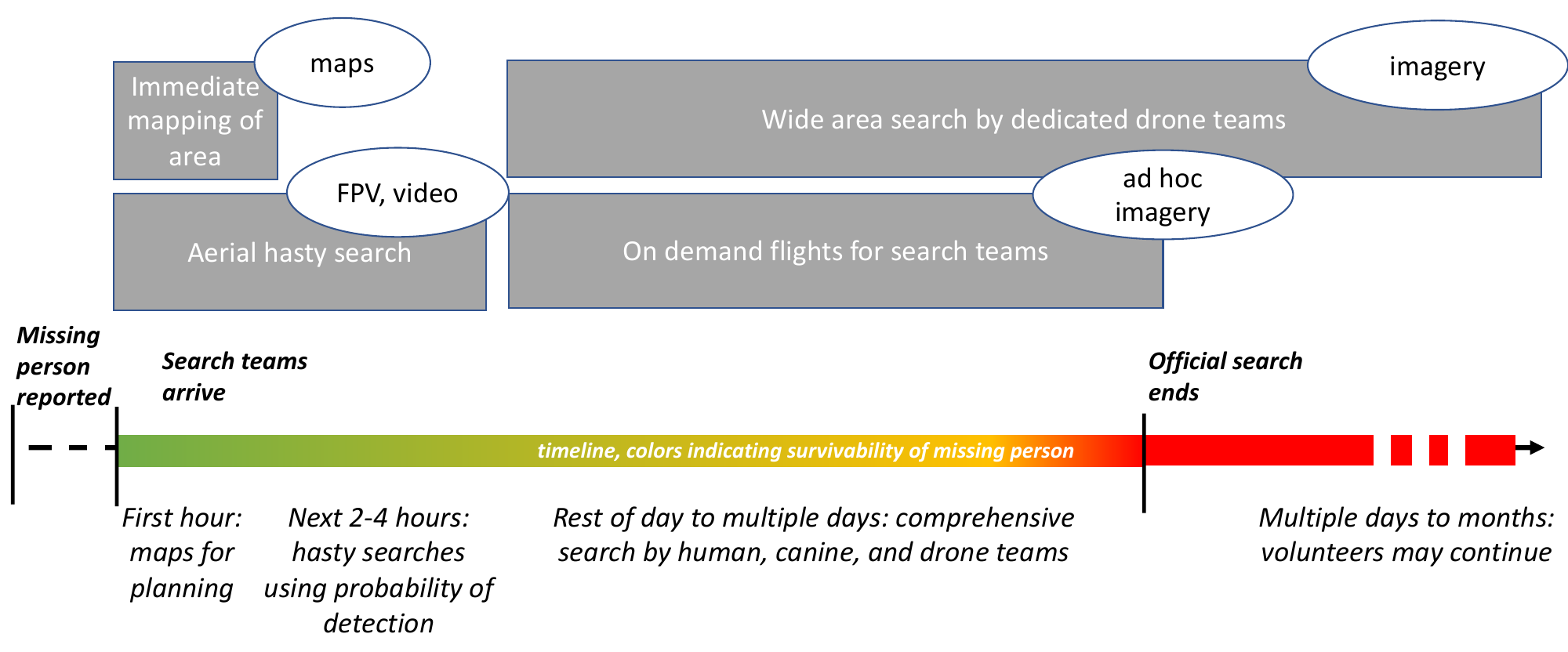}}
\caption{Notional work process for drones at WiSAR incidents.}
\label{fig:wisar}
\end{figure*}

Figure ~\ref{fig:wisar} graphically captures the notional work process for how drones are used at WiSAR incidents and where CV/ML is appropriate. The work process was
developed by CRASAR based on participation in over two dozen incidents, a similar number of high fidelity exercises and training courses, and ongoing discussions with the rescue community. This notional work process is different from the mountain search and rescue (MSAR) work flow, which assumed aerial assets from the beginning of the search \cite{schomer:2022}. It should be emphasized that every incident is different and different drone teams and agencies may have different procedures; this process and timeline is a synthesis of an overall framework. The data from wide area searches offers the greatest opportunity for CV/ML techniques because of the nearly intractable number of images that would otherwise have to be manually inspected. 

When the search for a missing person is engaged, the agency having jurisdiction has to spend approximately an hour coordinating resources and planning what groups (ground searchers, canine, equine, and drone teams) are participating, and with what capabilities. Part of the planning requires dividing up the area to be searched. Unfortunately satellite imagery may be outdated. Thus, if a drone squad is available, they will often fly while the other assets are checking in order to quickly create a low resolution map. That map can help identify roads, game trails, other access points, and changes in vegetation that would influence the distribution of search resources. This map is generally not sufficient quality to identify a missing person unless they are in the open and wearing bright clothes. 

The incident command also may deploy assets, including drones, for a hasty search of the areas with the highest probability of the location of the missing person \cite{koester}. The aerial hasty search is usually a first person video search of areas with high visibility
such as following roads, pipelines, railroad and utility easements, open fields, and bodies of water. The first person video search is based on the expectation that the pilot will be able to see the missing person or signs of them, or will see them in a replay of the recorded video. If thermal sensors are available, the squads will typically use both electro-optical (EO) and thermal imaging, though thermal typically works best in mornings and evenings where there is a greater temperature contrast. Thermal imaging can be hard for humans to interpret and animals may also produce false positives.  Drone teams may use software packages such as Loc8\cite{loc8}, which perform color and thermal intensity segmentation. 

As the search continues, the emphasis turns to conducting a thorough, systematic search. The area is generally divided into sectors to be completely searched, or researched. Drone quads may be embedded with ground teams to provide them aerial imagery on demand. This is particularly helpful when searchers encounter fences, dense vegetation, or a wide field and need to see what is beyond. Very few images or videos are collected in this case.  Drone squads may be assigned a sector that they are solely responsible for. In this case, they would typically fly an autonomous survey pattern to gather images that completely cover the polygon in question. The survey software is the same as for collecting imagery for constructing orthomosaics is used, popular packages are DroneDeploy and Pix4Dcollect. These packages assume that the images are taken with the camera facing down (nadir) and allow the pilot to specify the altitude and overlap of images in that polygon.  In addition, the drones may fly first person view to attempt to peer into a forest canopy or around a lake with an oblique camera angle. If the missing person is not found, the authorities will stop the search. At that point, private teams may continue the search. These  teams will generally survey, or re-survey, the area of interest with the mapping packages and some first person view flights.

Depending on the altitude, overlap in images, and area to be covered, a wide area survey may produce between 1,000 and 50,000 images. Each image has to be carefully examined for signs of the missing person as the person may be under a tree, trapped or at an unusual pose due to broken bones, or covered in mud or leaves. Thermal is rarely used after the missing person is presumed deceased; the victim would present a relatively small heat signature for a short tie during the initial body decay process. 

The images are usually inspected by trained members of the search team, referred to as squinting. The squinters typically work at the base of operations where they can use a larger monitor inside a tent or command trailer. Wireless communications and upload rates to speed up the data transfer process described in \cite{schomer:2022} are not an issue for the post-incident search since the private search activities, which are the focus of this paper, do not have the time pressure of the agency search, as the missing person is now presumed to be deceased.  If the squinter finds signs of a missing person in the image, ground search teams are given the GPS location. This is another advantage of nadir images, as the GPS location of the image is the center of the image and the distance to other pixels can be estimated. Squinting is time consuming and subject to human error as squinters may be fatigued from the search and the task requires a great deal of concentration, meaning they may miss objects or have a tendency not to examine the entire image.  

\section{Wu-Murad Missing Person Search}


Patricia Wu-Murad is a 60-year-old female hiker who went missing on April 10, 2023, while covering an 11 mile mountainous portion of the Komado Komo trail in southern Nara, Japan.  The formal search by the local authorities began on April 11 and involved 26 police officers and helicopters \cite{dailyMailWuMurad}. Canine teams from the Japan Rescue Association were also used through out the search, but did not get indications. Drones were considered but not used for the initial search due to the the dense tree canopy reducing likelihood of success. The search was terminated on April 13, 2023, without finding Wu-Murad, as the teams had searched as completely as possible given the dangerously steep terrain and  the probability of her survival was negligible. 
%
From April 14 to May 30, 2023, the family engaged private ground and drone teams from the United States, who expanded the search area. The drone teams collected 98.9 GB of drone imagery and video between April 23 and May 15 and despite the use of Loc8 and squinting, they had no success in identifying any signs of Wu-Murad. On May 3, the family requested CRASAR to assist. CRASAR applied three different CV/ML methods, and identified only four images where there might be signs of a person. On May 30, 2023, the private search was terminated by the family.

\subsection{CV/ML Methods Used}

CRASAR was provided with
128 high resolution photos and 243 high resolution videos. The high resolution video was sampled, producing a total of 18,759 images. As described in \cite{manzini:ICCV}, three CV/ML approaches were used:
\begin{itemize}
\item EfficientDET, a supervised learning model trained on HERIDAL dataset. This reduced the 18,759 images to 1377 images, which were manually inspected. Four of the images were passed to search teams. 
\item Unsupervised RX spectral classifier used in the 2015 Blanco River Flood \cite{proft:2015}. This reduced the number of images 
 to 5234 images, but manual inspection yielded no images for follow up. 
\item YOLOv5 was abandoned due to poor validation set performance.
\end{itemize}

\subsection{Data Acquisition in the Wu-Murad Search}

 \begin{figure}[htbp]
\centerline{\includegraphics[width=3.8truein]{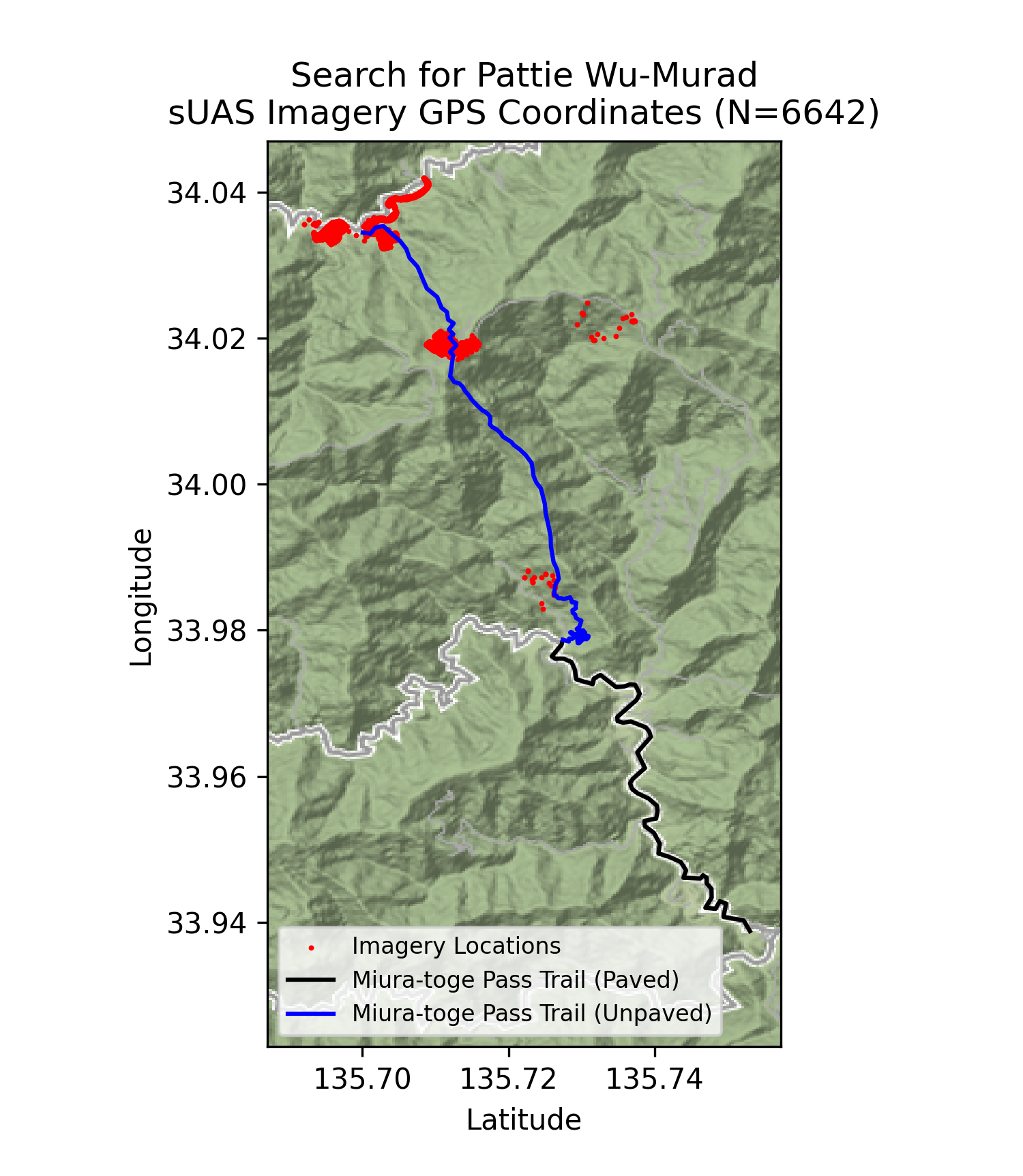}}
\caption{Areas searched with drones in red, segment of the trail that Wu-Murad was hiking is in black and blue. Some areas are missing due to lack of georeferencing data. }
\label{fig:searchArea}
\end{figure}

The Wu-Murad incident illustrates the timing  of the notional work process for drones. The  activities were consistent with the general timeline: the person was reported missing, the agency-led search lasted for three days, and then private teams took over. 

The incident does show some differences with the activities in the notional work process for wide area search, most notably the absence of a wide area complete coverage search by the  private drone teams.  
Figure ~\ref{fig:searchArea} shows the location of flights from the subset of data which was geotagged, and suggests that the drones did not perform a comprehensive complete coverage search of the area, though this cannot be confirmed. The low number of images also indicates that only a small portion of the area was searched.   The reason for the lack of a comprehensive search is unknown, but it should be noted that much of the area is steep vertical cliffs which would be searched in a manner similar to inspecting a building with a drone, not with a complete comprehensive search and squinting. 

\begin{figure}[htbp]
\includegraphics[width=0.47\textwidth]{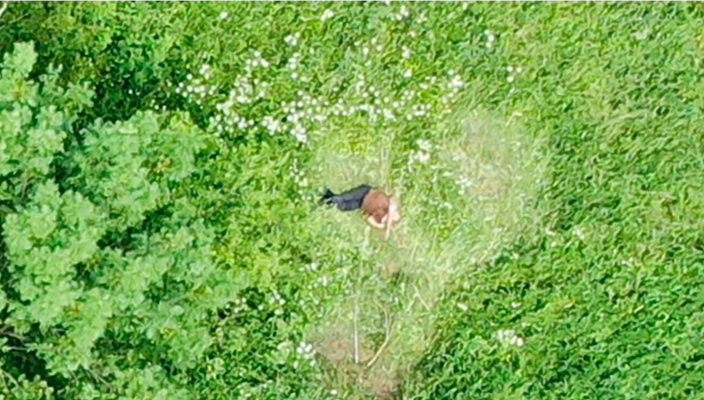}
\vspace{0.075in}
\centerline{a.}

\includegraphics[width=0.47\textwidth]{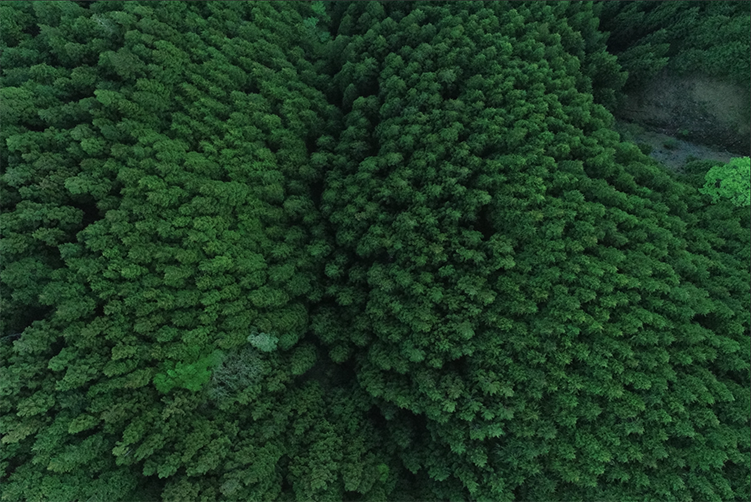}
\vspace{0.075in}
\centerline{b.}

\includegraphics[width=0.47\textwidth]{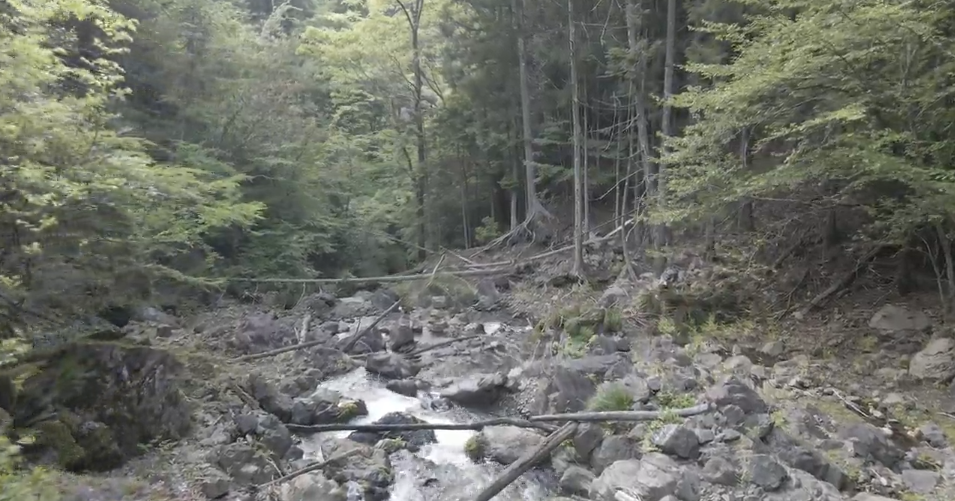}
\centerline{c.}
\caption{a) Drone image of missing person found successfully search reported in \cite{Tunsio:2022}, b) typical nadir image in the Wu-Murad search, c) typical oblique image in the Wu-Murad search.}
\label{fig:comparison}
\end{figure}

The Wu-Murad dataset is both challenging and realistic due to the wilderness terrain.  
The search area was not well suited for aerial search because it was comprised of 
mountainous terrain covered in forests with a dense tree canopy. 
Figure ~\ref{fig:comparison} shows the considerable visual difference between the successful use in Poland \cite{Niedzielski:2021,Tunsio:2022} and the environment in Japan. In Poland, the missing hiker was ambulatory, in a normal posture, and in an open area. In the Wu-Murad search, the missing hiker was presumed injured and thus may not be present a ``normal" silhouette, and would likely be covered in dirt, twigs, and leaves. The missing persons in the Blanco River floods \cite{proft:2015} were even more challenging to recognize in images, as they were presumed to be swept one to three miles down river, covered in mud, and tangled in debris. 

The Wu-Murad search also highlights the shortcomings of the dataset for post-processing by any algorithm, either photogrammetics or CV/ML. The data was a heterogeneous mixture of oblique and nadir images and video taken at
 different times of day and from different altitudes, as would be expected from the notional work process. The CV/ML algorithms 
 described in \cite{manzini:ICCV} expect the size of the missing person in the image
 to be roughly the same size as the bounding box of the person in the training images. In the case of the HERIDAL
 dataset, the average bounding box was a square size with width and height of 64 pixels and a standard deviation of 23 pixels \cite{maruvsic2018region}. This means that the altitude should be set for the particular camera to capture images with this resolution. In addition, the camera angles are expected to be nadir, not oblique; there is no guarantee that a CV/ML model trained on nadir images will 
 successfully identify targets from oblique camera views. 
 

\section{Facilitating Acquisition for CV/ML}

The work process for drones and the lessons learned from the Wu-Murad search leads to five recommendations for facilitating acquisition of images that are more likely to be effectively used by CV/ML models. 

First, imagery should be high resolution EO georeferenced camera data, not video or thermal. 
The use of still images versus video is essential for effectively utilizing CV/ML because all datasets for WiSAR tasks are based upon still images, video will contain imagery with unnecessarily high overlap percentages (in the form of adjacent frames) that must be filtered out to ensure reasonable processing times, and sUAS typically capture still images at a higher resolution than video. 
Georeferencing is also critical because the purpose of applying CV/ML is not just to find a missing person but also to give the location. Fortunately, almost all drone cameras are georeferenced. Video frames can be georeferenced if 
the drone uses a captioning system (.srt file) that associates the GPS location with the sampled image, but not all drones have this feature. Therefore it is incumbent on the drone team to use platforms which have captioning and to make sure the feature is turned on. 
While the WiSARD dataset \cite{broyles2022wisard} recently introduced annotated thermal imagery training data for person detection, WiSAR tasks involving thermal imagery remains an open problem for CV/ML systems. In general, the use of thermal imaging becomes less valuable as the search extends into weeks. In addition, not every inexpensive drone has thermal camera and if they do, it is typically low resolution. 
Other sensors such as lidar, hyperspectral, and synthetic aperture radar have not been proven to work for these situations and require specialized drones.

Second, imagery should be collected with an autonomous survey planning package to guarantee
complete coverage of a specified area. CV/ML cannot find a missing person if there is no image.  While complete coverage is important for the CV/ML model to be applied to an area of interest, the automated collection should also consider the manual
squinting step to confirm the target in the image. This can be facilitated by selecting the degree of overlap of images. An
overlap of 60\% should be sufficient to ensure that each pixel appears in at least two images, providing a limited set of alternate views to help squinters confirm  possible
targets; this is similar to the "synthetic hovering'' concept proposed in \cite{Bashyam:2019}. 60\%  is lower than the 70-75\% recommended for generating hyper-accurate orthomosaic maps; however, mapping is not the objective and the lower overlap conserves drone batteries. At the same time, higher percentages of overlap are recommended if there are many vertical obstacles, such as trees, that obstruct the view of the ground near the edges of the frame.

Third, the camera angle should be nadir, straight down. While this viewpoint is harder for humans to manually interpret imagery, it is the camera angle that existing CV/ML models are trained with. A nadir viewpoint also simplifies estimating the GPS location of the target for the ground teams to reach. 
Autonomous survey software will typically default to a nadir camera angle. 

Fourth, it is desirable to collect imagery mid-day to reduce shadows, which confound target identification.
Although restricting flying to mid-day is usually not practical due to time and resource constraints for flying, minimizing shadows should improve CV/ML identification as shadows obscure color, shape, and other other features of the missing person. 

\begin{figure}[htbp]
\centerline{\includegraphics[width=0.47\textwidth]{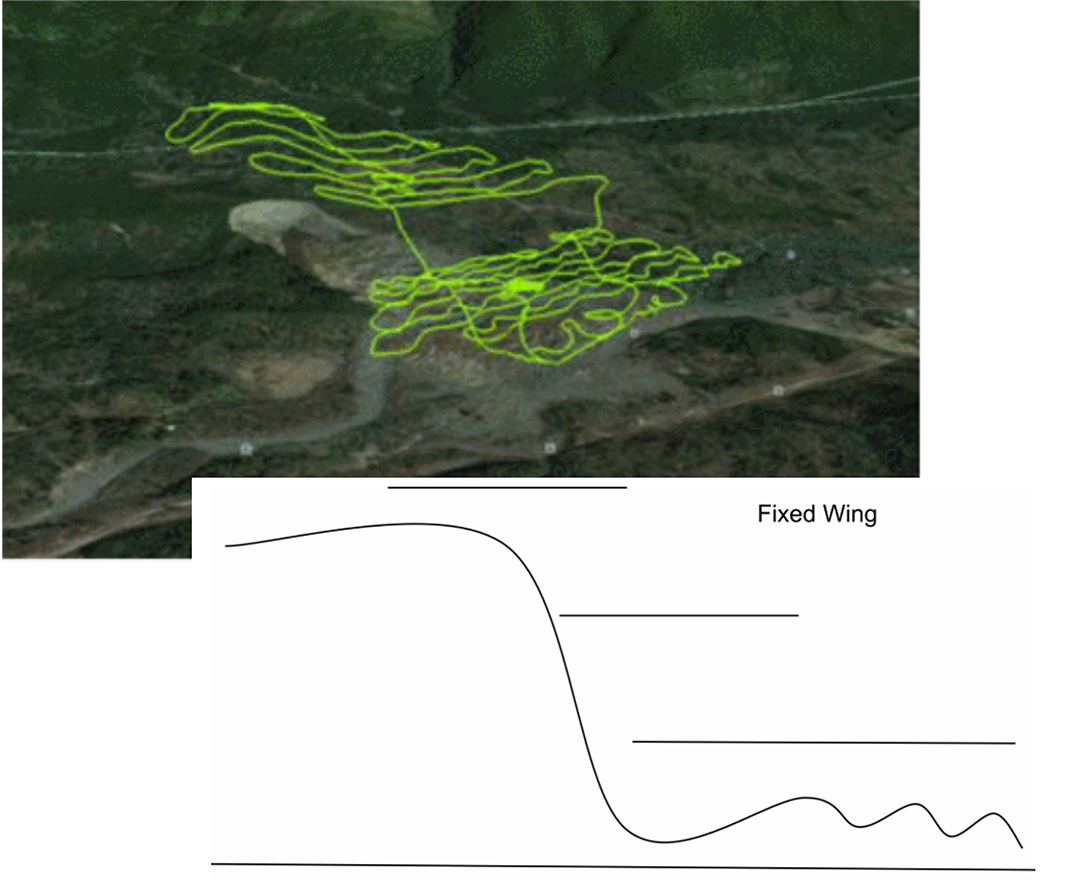}}

\caption{``Stair stepping'' strategy for acquiring images with the same above ground level altitude in mountainous terrain after \cite{duncan:osoflights}.}
\label{fig:stairStepping}
\end{figure}

Fifth, the altitude should be set as a function of the average size of the targets in the CV/ML model, otherwise it should be the lowest altitude reasonable in order to guarantee coverage and consistent resolution. In mountainous terrain, one strategy for acquiring images from the same altitude is ``stair stepping'' approach developed for the 2014 Oso Mudslide response \cite{duncan:osoflights} and later in river terrace mapping \cite{Li:2019}.  As shown in Figure ~\ref{fig:stairStepping}, hilly terrain is divided into polygons of roughly level terrain. Each polygon can be flown at the same altitude above ground level though with notable differences between ``steps" in terms of altitude above mean sea level. Given that the terrain may only be roughly level and that trees do not grow to the same height, the pilot may have to fly to the tallest tree to determine the lowest safe altitude \cite{duncan:osoflights}. However, stair-stepping will not work on cliff faces and steep drop-offs, and there may be other regions of an area that existing CV/ML models based on nadir imagery are not applicable.

\section{Conclusions}

As drones are increasingly used for missing person search and rescue, CV/ML will have a greater role to play. The extremely challenging Wu-Murad search serves as a case study of how drone data acquisition impacts the successful application CV/ML to wilderness missing person search, especially for the data-intensive wide area search activity of the larger search work process. The 2023 search illustrates that ad hoc drone data collection is unlikely to meet the expectations of CV/ML post-processing models. Although the CV/ML models and datasets could be greatly improved \cite{manzini:ICCV}, data collection can be improved to gather images more compatible with the expectations and constraints of these models. Fortunately, drone teams can adapt flight strategies to collect more suitable data and platform designers can ensure that minimal functionality is provided in new drones.

Drone teams conducting wide area searches may not know if the data they collect is going to be used for CV/ML post-processing instead of manual squinting, but they can improve the likelihood that the data will be compatible with automated detection. One way is to use automated collection software to collect a set of imagery that  completely covers the area of interest at a consistent altitude above ground level. The images should be geo-tagged, high-resolution EO nadir images, not video or thermal (which are valuable for other search activities and times in the search). If possible, flights should be conducted at mid-day to reduce shadows.  If the drone teams do expect to use a specific CV/ML model, they can use knowledge about that specific model to optimize the altitude or other factors.  

The gap between data collection and CV/ML post-processing offers opportunities for research and development. Advances in automating planning for ``stair stepping'' in mountainous terrain to collect data in patches with a consistent altitude above ground \cite{duncan:osoflights,Li:2019} would be valuable. 
There is no indication that a specialized drone is needed for CV/ML imagery collection in wilderness search and rescue, as any drone capable of taking high resolution geo-tagged images and executing path planning software should work. Other sensors such as thermal and functionality such as delivery may be of value to the larger search and rescue enterprise, but have limited value for CV/ML. 
In terms of multi-robot coordination, any coordination software has to support all models of drones, as there is no guarantee that agencies and volunteer teams will have the same drones. The software should be easy to install in the field and use with minimum of training, and be free or low-cost, otherwise it is unlikely to be adopted by volunteer teams. While multi-robot coordination schemes typically focus on assigning tasks, coordination software that also enforced a consistent data labeling scheme from all drones and ensured quality control would be useful. 

 At the time of writing Patricia Wu-Murad remains missing; the data from the Wu-Murad search and CV/ML models is available at \url{https://github.com/CRASAR/WiSAR/}. We welcome collaboration and experimentation. 

\section*{Acknowledgments}

The authors would like to thank the Wu-Murad family and Christopher Berquist for making the datasets available and we continue to express our sympathy for the family. Satoshi Tadokoro and Kazunori Ohno kindly provided insights into the search through their Cyber Canine Rescue connections. The notional work flow was developed in collaboration with David Merrick, Justin Adams, and Jarrett Broder. 

\bibliographystyle{IEEEtran}
\bibliography{Bib/main}
\end{document}